\documentclass[letterpaper, 10pt, conference]{ieeeconf}

\IEEEoverridecommandlockouts
\usepackage[T1]{fontenc}
\usepackage{graphicx}
\graphicspath{{figures/}}
\usepackage{amsmath,amssymb,amsfonts,bm}
\usepackage{mathtools}
\usepackage{booktabs,multirow}
\usepackage[caption=false,font=footnotesize]{subfig}
\usepackage[table]{xcolor}
\usepackage{cite}
\usepackage{url}
\usepackage{xspace}
\usepackage{cite}
\makeatletter\let\NAT@parse\undefined\makeatother
\usepackage[hidelinks]{hyperref}
\hypersetup{pdfauthor={Bingao Chen, Lihan Zha, C. Karen Liu, Guy Tevet}, pdftitle={FoveaVLA: Foveation Actions Replace the Wrist Camera in VLAs}, pdfsubject={}, pdfkeywords={}}

\ifdefined\FINAL
  \newcommand{\DraftOnly}[1]{}
\else
  \newcommand{\DraftOnly}[1]{#1}
\fi

\usepackage{capt-of}

\usepackage{xspace}

\newcommand{\modelname}{FoveaVLA\xspace}
\newcommand{\pifive}{$\pi_{0.5}$\xspace}

\newcommand{\baselineBlackout}{\pifive-Blackout\xspace}
\newcommand{\baselineECoT}{ECoT-inspired\xspace}

\newcommand{\altDualRGB}{\modelname{}-Dual-RGB\xspace}
\newcommand{\ablSingleRGB}{\modelname{}-Single-RGB\xspace}

\newcommand{\val}[2]{#1\,{\scriptsize\textcolor{black!55}{$\pm$#2}}}

\definecolor{gtcolor}{RGB}{0, 90, 200}
\definecolor{bccolor}{RGB}{200, 30, 30}
\definecolor{klcolor}{RGB}{0, 140, 60}
\definecolor{lzcolor}{RGB}{150, 0, 180}

\ifdefined\FINAL
     \newcommand{\gtc}[1]{}
     \newcommand{\bcc}[1]{}
     \newcommand{\klc}[1]{}
     \newcommand{\lzc}[1]{}
\else
  
  \newcommand{\gtc}[1]{\textcolor{gtcolor}{[GT: #1]}}
  
  \newcommand{\bcc}[1]{\textcolor{bccolor}{[BC: #1]}}

  \newcommand{\lzc}[1]{\textcolor{lzcolor}{[LZ: #1]}}
\fi

\title{
\LARGE \bf \modelname: Foveation Actions Replace the Wrist Camera in VLAs
}

\author{%
\authorblockN{Bingao Chen$^{1}$, Lihan Zha$^{2}$, C. Karen Liu$^{1,\dagger}$, Guy Tevet$^{1,\dagger}$%
\thanks{$^{\dagger}$Equal advising.}%
\thanks{This work was done during Bingao Chen's internship at Stanford University. Bingao is now with Tsinghua University.}}%
\authorblockA{$^{1}$Stanford University \qquad $^{2}$Princeton University\\
\href{https://tml.stanford.edu/FoveaVLA/}{\textcolor{blue}{\texttt{tml.stanford.edu/FoveaVLA}}}}%
}

\IEEEaftertitletext{%
  \vspace{-1.5\baselineskip}%
  \noindent\begin{minipage}{\textwidth}%
    \centering
    \includegraphics[width=\textwidth]{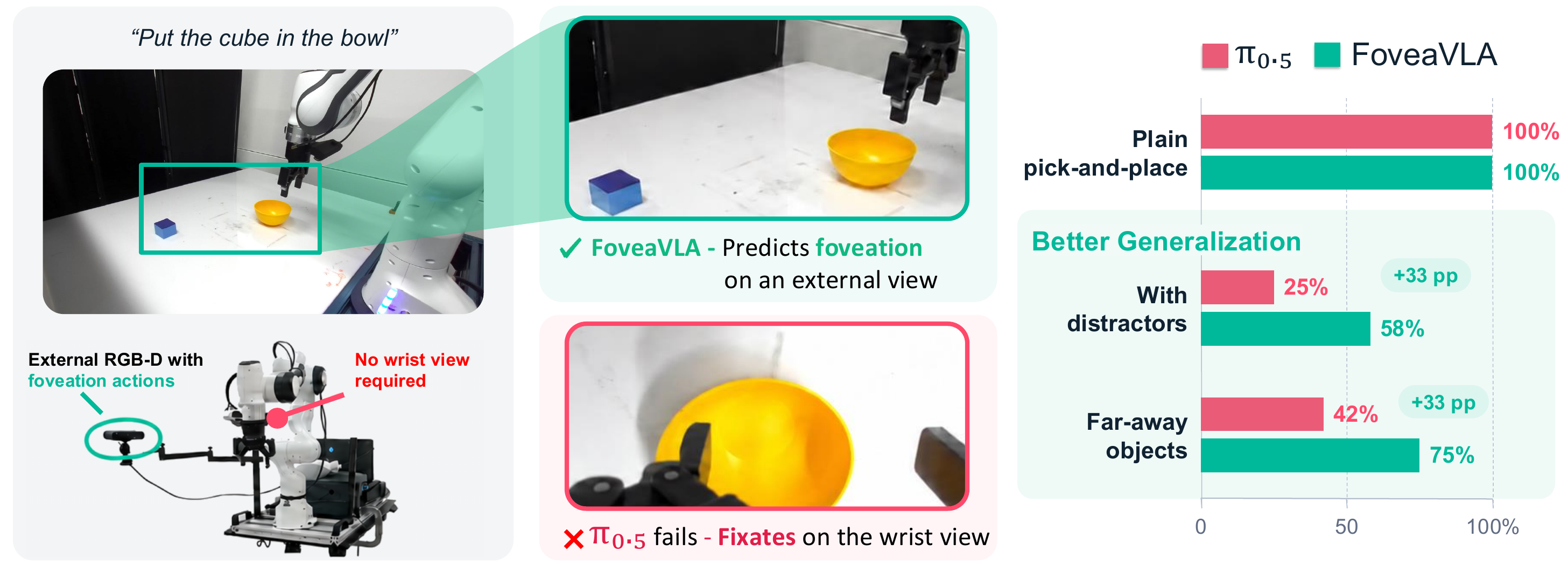}
    \captionof{figure}{
\modelname replaces the wrist camera with learned foveation actions: it predicts the next interaction region and observes it at high resolution. The wrist-equipped \pifive matches it when the target is already in the wrist view, but degrades with distractors and far-away objects, where \modelname's foveal crop keeps the relevant detail in sight.
    }
    \label{fig:teaser}
  \end{minipage}%
  \vspace{0.5\baselineskip}%
}

\begin{document}

\maketitle
\thispagestyle{empty}
\pagestyle{empty}

\begin{abstract}
We present \modelname, a Vision-Language-Action (VLA) model that replaces the wrist camera with learned foveation actions. Wrist cameras provide close-up detail but restrict the view to the end-effector's surroundings.
External cameras see the whole scene, but the policy receives them at low resolution, so the fine detail needed for manipulation is lost.
\modelname recovers this detail by predicting, alongside each action chunk, the interaction region to observe at the next policy query. The policy then receives crops of these regions, taken from the original-resolution images, together with the full views. We build \modelname on \pifive in two configurations, Single and Dual, using one or two external stereo cameras with full and foveal RGB and depth inputs, and post-train both on DROID with RoboInter interaction annotations. Across four real-world task suites, \modelname{}-Single achieves $70.8$\% success with one external RGB-D view, and \modelname{}-Dual achieves $77.1$\%, versus $52.1$\% for the wrist-equipped \pifive. Both match the baseline on plain pick-and-place and improve success with distractors, cloth manipulation, and far-away objects.
\end{abstract}

\section{Introduction}

Recent progress in robot manipulation has been driven by large models trained on diverse robot datasets~\cite{o2024open,kim2024openvla,black2024pi_0}. Most of these models take a wrist camera as input~\cite{zhao2023learning,khazatsky2024droid,chi2024universal}, which provides close-up observations that learned policies often depend on~\cite{hsu2022vision,fei2025libero,huang2026selective}. This close-up view comes at the cost of scene coverage: the camera moves with the end-effector, so distant targets and surrounding context can lie outside its view, more so when manipulation spans a larger workspace or involves a mobile robot. The wrist camera also adds hardware and creates an observation mismatch with human demonstrations recorded without a wrist view~\cite{kareer2025egomimic,lepert2025phantomtrainingrobots}. Can a policy get the close-up detail it needs from external observations alone, so that the robot no longer needs a wrist camera?

External cameras record high-resolution images, but resizing a whole scene to the policy's low input resolution reduces the detail available for the gripper and object, and uniformly increasing the input resolution spends computation on background regions. External views also lack a consistent geometric frame: a wrist camera maintains a fixed pose relative to the end-effector, whereas an external view depends on the camera's placement relative to the robot. Prior work reports lower success for policies learned without wrist observations~\cite{robomimic2021,hsu2022vision}. We observe a similar gap on our four real-world task suites: \pifive post-trained on two external RGB views alone achieves $25.0$\% success, compared with $52.1$\% with the wrist view.

Existing approaches obtain close-up observations through additional sensing or constructed viewpoints. Generative methods synthesize wrist images from external observations~\cite{qian2025wristworld,ding2025imagination}. Others render virtual views from 3D scene representations~\cite{Liu_2026_CVPR,goyal2024rvt}, or move a camera on a dedicated arm or neck~\cite{chuang2025active,xiong2025vision}. Human-gaze-guided foveation is closer to our goal of selecting detail within an existing image~\cite{kim2021gaze,chuang2025look}, but relies on gaze recorded during data collection. We instead learn where to crop from existing robot demonstrations, supervised by their interaction annotations.

Human vision allocates detail selectively. Peripheral vision provides scene context while central vision resolves the detail needed to grasp an object~\cite{sivak1990integration}, and gaze reaches the next contact point before the hand does~\cite{johansson2001eye}. We take the same approach in \modelname, which replaces the wrist camera with a learned foveation action that selects where to look next. Cropping the predicted interaction region from the original-resolution image and resizing it to the same $224\times224$ encoder input devotes many more pixels to the gripper and object than the resized full view does. Feeding this crop alongside the full view gives the policy both the close-up detail it needs for manipulation and the layout of the whole scene.

We build \modelname on \pifive~\cite{black2025pi_}. From the current observation, it generates a robot action chunk and predicts one bounding box per external view for the interaction region at the next policy query. The predicted boxes crop the newly captured images, and the crops are added to the next policy input alongside the full views. We derive supervision from RoboInter annotations~\cite{li2026robointer} on DROID demonstrations~\cite{khazatsky2024droid}, constructing a bounding box around the components involved in each annotated skill, such as the gripper, object, and target region. Annotations at the next query provide targets for future-box prediction. We instantiate the method as \modelname{}-Single and \modelname{}-Dual, using one and two external stereo cameras, respectively, with full and foveal RGB images and aligned depth estimated from each stereo pair. Both operate without a wrist camera; Dual adds a second external viewpoint to the same foveation mechanism.

We post-train both configurations on DROID and evaluate them on the four real-world task suites. \modelname{}-Single achieves $70.8$\% success using one external RGB-D view, exceeding the wrist-equipped \pifive at $52.1$\%, and \modelname{}-Dual achieves $77.1$\%. The largest gains are on pick-and-place with distractors, where Single raises success from $25.0$\% to $58.3$\%, and on moving far objects, from $41.7$\% to $75.0$\%. On MolmoSpaces~\cite{kim2026molmospaces}, \modelname{}-Dual achieves $22.4$\% average zero-shot success versus $17.8$\% for \pifive-DROID~\cite{openpi}, outperforming it on seven of nine tasks.

\section{Related Work}

\textbf{Wrist cameras in robot manipulation.}
Eye-in-hand cameras date back to visual servoing~\cite{chaumette2006visual} and were adopted in learning-based manipulation after empirical studies showed their benefit~\cite{robomimic2021}. Hsu et al. argued that the partial observability of the hand-centric view is an inductive bias that aids generalization~\cite{hsu2022vision}. Wrist cameras are now standard in bimanual hardware~\cite{zhao2023learning}, handheld collection devices~\cite{chi2024universal}, datasets and benchmarks~\cite{khazatsky2024droid,NEURIPS2023_8c3c6668,kim2026molmospaces}, and generalist policies~\cite{black2025pi_,liu2025rdt,kim2025fine}.
Wrist-only observations can lack the scene context needed for manipulation, motivating a wide-angle fisheye lens in UMI~\cite{chi2024universal} and an additional external view in MV-UMI~\cite{11594803}.
Policies also come to depend on this view: LIBERO-Plus finds that multi-view VLAs retain much of their performance with the external view masked~\cite{fei2025libero}, and other studies find policies including \pifive more sensitive to masking or perturbing the wrist view~\cite{huang2026selective,heo2026anycamvla,yang2026robolab}.
\modelname keeps the close-up that makes the wrist view valuable, without mounting a camera on the hand.

\textbf{Constructed viewpoints and active vision.}
WristWorld generates wrist-view videos from anchor views~\cite{qian2025wristworld}, while inference-time view synthesis supplies in-hand images to a trained policy~\cite{ding2025imagination}. WARPED renders wrist-aligned training observations from egocentric human videos but retains a physical wrist camera during execution~\cite{freeman2026warped}. Others re-render the scene from virtual viewpoints using 3D scene representations, zooming coarse to fine on point clouds~\cite{goyal2024rvt} or selecting virtual views and applying 3D zoom within a VLA~\cite{Liu_2026_CVPR}. Physical active-vision systems instead control a camera using demonstrations~\cite{chuang2025active,xiong2025vision,yu2025egomi} or task reward~\cite{kerr2025eye}. EyeVLA predicts pan, tilt, and zoom for a physical PTZ camera~\cite{yang2025look}. Such systems require camera actuation and supervision or rewards for camera control.
\modelname uses fixed cameras and selects regions within the images they already capture.

\textbf{Learned visual attention and foveation.}
Human vision resolves fine detail only in a small central region, and the eyes move this region to the grasp point before the hand arrives~\cite{strasburger2011peripheral,sivak1990integration,johansson2001eye}.
Attention models learn glimpse actions~\cite{mnih2014recurrent}, and foveation also aids recognition~\cite{blauch2026fovi} and gaze-conditioned generation~\cite{chao2026foveateddiffusion}. VLMs retrieve local evidence through crops~\cite{10657571,zheng2026deepeyes,shabtay2026look,min2026foveated} to answer questions about a given image, whereas we use predicted regions to select the next robot observation. In manipulation, foveation has been supervised by human gaze, as a foveal crop fed to the policy~\cite{kim2021gaze} or gaze co-predicted with actions~\cite{chuang2025look}, a signal recorded with eye trackers during collection and not provided by DROID. AtVLA corrects visual attention with register tokens and, when uncertain, crops a region chosen by attention rollout~\cite{cui2026look}. Another approach trains a policy that chooses which regions a sensor captures for frozen downstream models, including a manipulation policy~\cite{10.1145/3799902.3811218}. \modelname instead decodes a future interaction box alongside each action chunk. Prior VLAs also use localization supervision~\cite{black2025pi_}, spatial traces~\cite{lee2025molmoact,zawalski2024robotic}, or visual and textual affordances~\cite{li2025coa}, but only as auxiliary targets that help action prediction.
In \modelname, the predicted box determines the crop the policy receives next, and its supervision comes from existing large-scale interaction annotations~\cite{li2026robointer}.

\section{Preliminaries}
\label{sec:prelim}

\textbf{Vision-Language-Action models.}
A VLA is a robot policy $\pi_\theta(A_t \mid o_t, \ell)$ that conditions on the current observation $o_t$, comprising camera images and the robot state $q_t$, and a language instruction $\ell$. It outputs an action chunk $A_t=(a_{t+1},\ldots,a_{t+H})$ spanning a prediction horizon of $H$ control steps. In receding-horizon execution, the controller can execute only the first $w\leq H$ actions before replanning. A common recipe starts from a pretrained vision-language model, adds an action head, and trains the resulting policy by imitation on robot demonstrations~\cite{kim2024openvla,black2024pi_0}. The visual encoder converts camera images into patch tokens, which the language-model backbone processes together with the instruction and robot state to form the representation used for action generation.

\textbf{Backbone and action generation.}
We build on \pifive~\cite{black2025pi_}. Its PaliGemma backbone~\cite{beyer2024paligemma} combines a SigLIP visual encoder with a Gemma language model, while a separate action expert generates continuous action chunks through flow matching, conditioned on the observation representation. PaliGemma's vocabulary also includes location tokens for image coordinates in detection and grounding. The \pifive base checkpoint does not emit these tokens at inference; we reuse them to represent foveation actions in Sec.~\ref{sec:foveation}.

\begin{figure*}[t!]
  \centering
  \includegraphics[width=\textwidth]{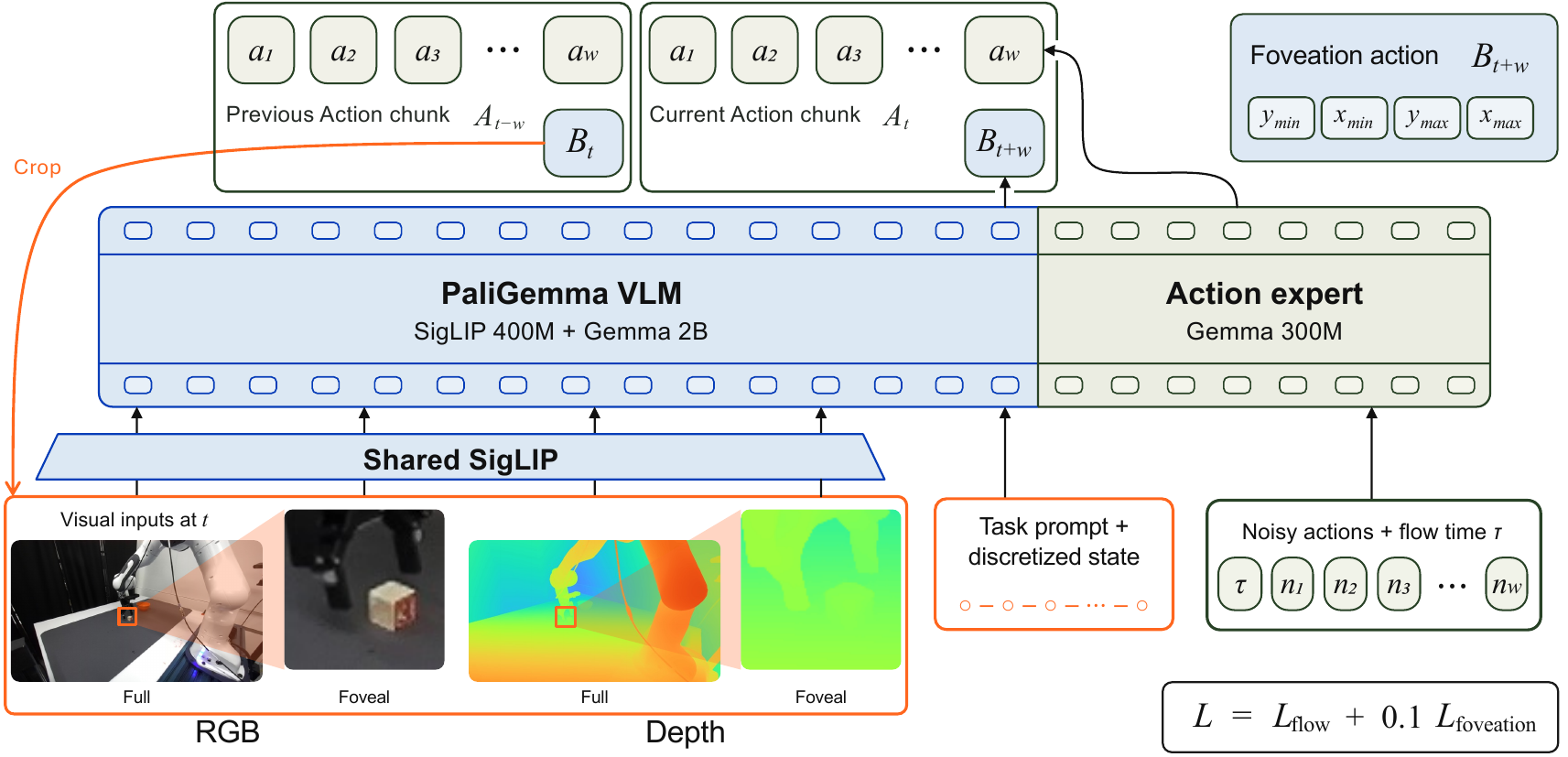}
  \caption{
  \textbf{\modelname{} overview (Single shown).} Full and foveal RGB images and aligned depth maps are encoded together with the task instruction and robot state. From this shared representation, the action expert generates an action chunk, while the language-model head autoregressively generates interaction boxes that select the foveal crops for the next observation. Dual adds the same four image inputs from a second external view and predicts one box per view.
  }
  \label{fig:training}
\end{figure*}

\section{Method}
\label{sec:method}

\modelname learns where to look alongside how to act (Fig.~\ref{fig:training}). From the full and foveal RGB-D observations, the robot state, and a task instruction, it predicts an action chunk and the interaction region for the next observation. The predicted regions determine which crops of the next external images the policy receives. We post-train it from the \pifive base checkpoint on DROID with RoboInter annotations. \modelname{}-Single uses one external stereo camera, and \modelname{}-Dual uses two.

\subsection{Full and Foveal Observations}
\label{sec:observations}

Let $N\in\{1,2\}$ be the number of external views for Single and Dual, respectively. Let $I_t^v$ be the source RGB image, at the camera's original resolution, from external view $v\in V = \{1,\ldots,N\}$ at time $t$, and let $b_t^v$ specify the region selected for detailed observation. The full view $G_t^v=\mathcal{R}(I_t^v)$ is the source image resized and padded to the encoder's low input resolution ($224\times224$ in our case). The foveal crop $F_t^v=\mathcal{C}(I_t^v,b_t^v)$ is a square crop around $b_t^v$, extracted at the source resolution and resized to the same low resolution. The full view preserves the scene layout, while the foveal crop spends the same number of pixels on the interaction region.

Each external camera is a stereo camera. We use Fast-FoundationStereo~\cite{wen2025fastfoundationstereo} to estimate disparity from its rectified image pair and convert it to metric depth using the focal length and stereo baseline. The depth map $D_t^v$ is aligned with $I_t^v$ and provides an explicit distance cue. We apply the same resizing and cropping to depth as to RGB, giving $D_{G,t}^v$ and $D_{F,t}^v$. Depth is mapped through a fixed logarithmic scale and colorized for the visual encoder; patches with insufficient valid depth are masked.
The resulting observation is
$
  o_t=\left(q_t,\left\{G_t^v,F_t^v,D_{G,t}^v,D_{F,t}^v\right\}_{v=1}^{N}\right),
$
where $q_t$ contains the robot's joint positions and gripper state. RGB and colorized depth use the same visual encoder and enter the transformer as separate sequences of visual tokens. The task instruction $\ell$ and discretized robot state complete the shared input prefix.

\subsection{Foveation Actions}
\label{sec:foveation}

We extend the VLA policy defined in Sec.~\ref{sec:prelim} to predict one bounding box per external view alongside the robot action chunk. Each box describes the interaction region at $t+w$, when the policy is queried next after executing the first $w$ actions. We predict $H=16$ actions and execute $w=8$ per query (Appendix~\ref{app:implementation}). We target $t+w$ because the gripper and objects move during the executed actions, and apply the predicted box to the image captured at that time.
Denoting $B_{t+w}=\{b_{t+w}^v\}_{v \in V}$
for the foveation actions, the joint policy factorizes as
\begin{equation}
  \pi_\theta(A_t,B_{t+w}\mid o_t,\ell)
  =\pi_\theta(A_t\mid o_t,\ell)\,
  p_\theta(B_{t+w}\mid o_t,\ell).
  \label{eq:policy}
\end{equation}
The two outputs share the observation representation, but are predicted by two independent heads: the action expert generates $A_t$, and the language-model head decodes $B_{t+w}$.

We initialize \modelname from the \pifive base checkpoint, retaining its backbone and flow-matching action expert. To predict $B_{t+w}$, we reuse the language-model output head to autoregressively decode the existing location tokens introduced in Sec.~\ref{sec:prelim}.

Each box is represented by four coordinates on the resized and padded full view, quantized into $1{,}024$ bins and serialized in $(y_{\min},x_{\min},y_{\max},x_{\max})$ order. The boxes form a sequence of $4N$ location tokens: four for Single and eight for Dual. The output vocabulary is restricted to location tokens. The action expert attends to the shared input prefix, while each box token also attends to the preceding box tokens. The two output branches are masked from each other.

Let $\hat b_{t+w}^v$ denote the predicted box in source-image coordinates. It determines the next foveal observation through
$  F_{t+w}^v
  =\mathcal{C}\!\left(I_{t+w}^v,\hat b_{t+w}^v\right),
$
with the same crop applied to the aligned depth map.

\subsection{Learning from Interaction Annotations}
\label{sec:training}

\begin{figure}[t]
  \centering
  \includegraphics[width=\columnwidth]{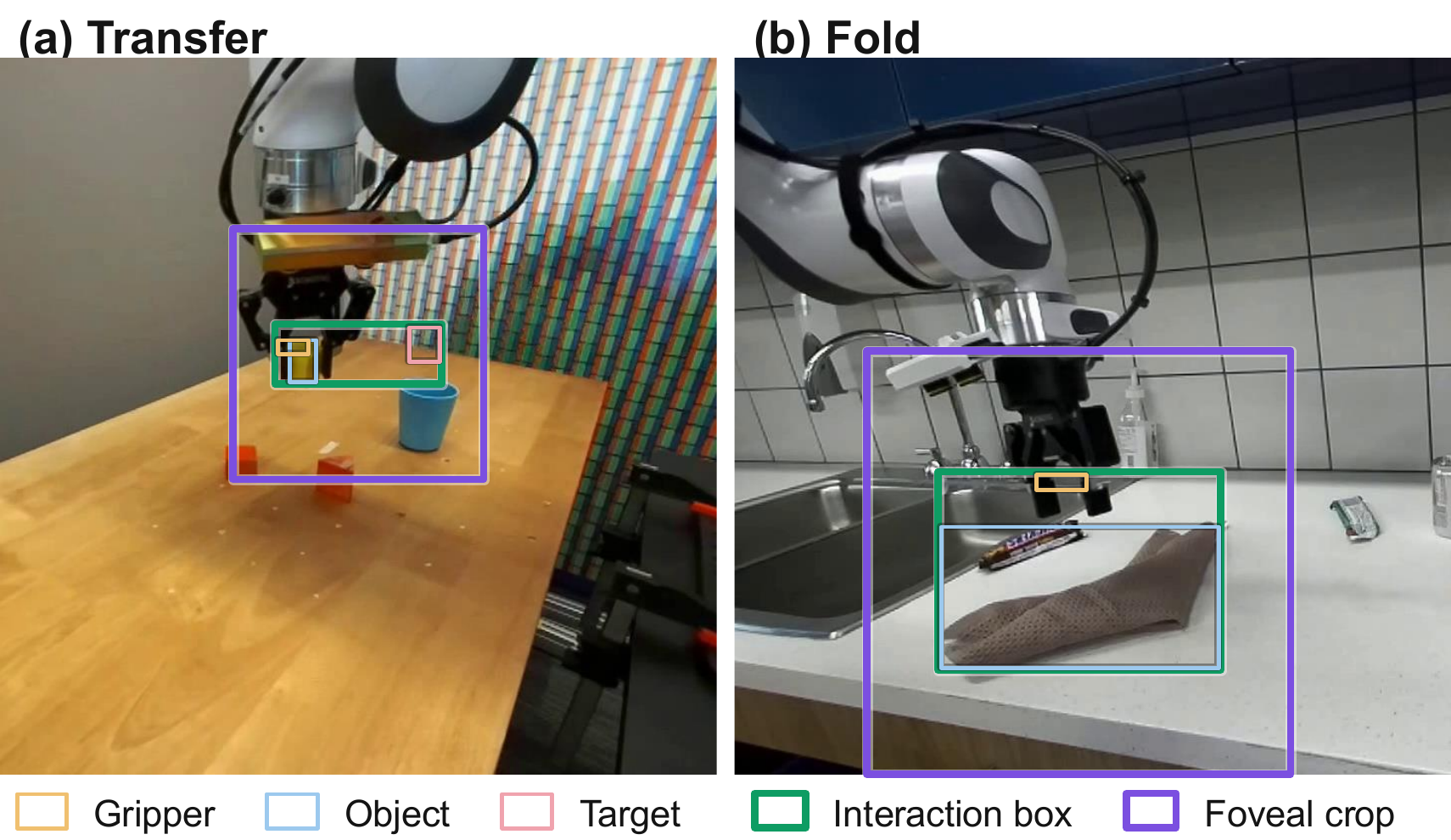}
  \caption{\textbf{Interaction boxes and foveal crops.} RoboInter annotations on DROID define interaction boxes (green) enclosing skill-relevant components; square foveal crops (purple) provide detailed visual inputs.}
  \label{fig:interaction_annotations}
\end{figure}

We derive foveation supervision from the RoboInter annotations associated with DROID demonstrations~\cite{li2026robointer,khazatsky2024droid}. For each annotated frame and external view, we construct one interaction box as the smallest axis-aligned rectangle enclosing the components required by the annotated primitive skill. Figure~\ref{fig:interaction_annotations} illustrates transfer and fold; both use the gripper, manipulated object, and annotated target region. Appendix~\ref{app:data} specifies the complete role mapping and preprocessing procedure. The interaction box defines the supervision target; the foveal crop is a square region enclosing this box, extracted as described in Sec.~\ref{sec:observations}.

A training example pairs the observation and demonstrated action chunk at $t$ with the interaction boxes at $t+w$. To improve robustness to imperfect predictions, we augment ground-truth crops at $t$ by perturbing their centers and scales while retaining the annotated region. Some crops are replaced by full views to match the initial query at inference. RGB and depth share the same crop geometry. Training examples require valid current and future annotations in the selected view for Single and in both views for Dual.

During post-training, we jointly optimize action generation and foveation prediction. For the action objective, we interpolate between a demonstrated chunk $A_t$ and Gaussian noise $\epsilon\sim\mathcal{N}(0,I)$ as $A_t^\tau=(1-\tau)A_t+\tau\epsilon$, where $\tau\in[0,1]$ is the flow time. The action expert $v_\theta$ learns the corresponding velocity field:
\begin{equation}
  \mathcal{L}_{\mathrm{flow}}
  =\mathbb{E}_{\mathcal{D},\tau,\epsilon}
  \left[\left\|v_\theta(A_t^\tau,o_t,\ell,\tau)-(\epsilon-A_t)\right\|^2\right],
  \label{eq:flow_loss}
\end{equation}
where the squared error is averaged over action entries.

For foveation prediction, let $z_{1:4N}^{*}$ be the location tokens for the $N$ annotated future boxes. We train their prediction with teacher forcing:
\begin{equation}
  \mathcal{L}_{\mathrm{foveation}}
  =-\mathbb{E}_{\mathcal{D},j}\left[
    \log p_\theta(z_j^{*}\mid o_t,\ell,z_{<j}^{*})
  \right],
  \label{eq:foveation_loss}
\end{equation}
where the expectation averages over training examples and uniformly over the $4N$ token positions. The total loss is
\begin{equation}
  \mathcal{L}=\mathcal{L}_{\mathrm{flow}}+\lambda\mathcal{L}_{\mathrm{foveation}},
  \label{eq:total_loss}
\end{equation}
with $\lambda=0.1$. Both losses update the shared backbone. We additionally apply dropout to the depth streams to train with partial depth availability.

\subsection{Closed-Loop Inference}
\label{sec:inference}

The initial foveal inputs are the full views. Subsequent foveal inputs use the preceding box predictions to crop the current RGB images and aligned depth maps, as described in Sec.~\ref{sec:foveation}. Each policy query encodes the observation prefix once for both outputs. The action expert generates the chunk by integrating the learned flow from noise to actions, and the chunk is returned immediately; the bounding box location tokens are then decoded greedily from the cached prefix while the robot executes the first $w$ actions, so the boxes are ready when the next frames arrive at $t+w$.

\section{Experiments}
\label{sec:experiments}

We evaluate learned foveation through five questions:
\begin{itemize}
  \item \textbf{(Q1)} Can learned foveation replace the wrist camera?
  \item \textbf{(Q2)} What do foveal crops add beyond box supervision?
  \item \textbf{(Q3)} What do depth and a second view add?
  \item \textbf{(Q4)} Does the policy transfer to unseen environments?
  \item \textbf{(Q5)} Can the policy predict where to look next?
\end{itemize}

\begin{figure*}[t]
  \centering
  \includegraphics[width=0.8\textwidth]{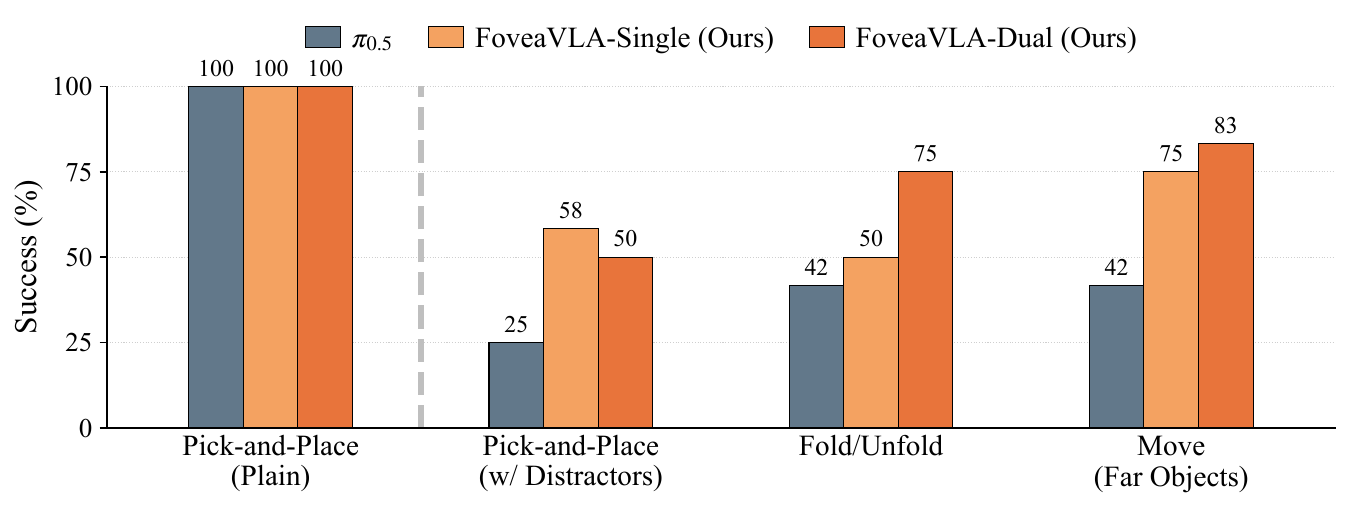}
  \caption{\textbf{Real-world task success} of wrist-equipped \pifive and the two \modelname{} variants, which use only external camera views. Each suite contains $12$ shared trials. Bar labels are rounded to the nearest integer; Table~\ref{tab:results} reports precise success rates, SEM, and all baselines.}
  \label{fig:main_task_results}
\end{figure*}

\subsection{Experimental Setup}
\label{sec:exp:setup}

\paragraph{Data and training}
We post-train \modelname{} from the \pifive base checkpoint on DROID~\cite{khazatsky2024droid}, comprising $\sim\!75$K teleoperated Franka Panda episodes with a Robotiq 2F-85 gripper, two external stereo cameras, and a wrist camera. Our models use only external views and RoboInter interaction-box supervision~\cite{li2026robointer} (Sec.~\ref{sec:training}), following the open-source training recipe~\cite{openpi}. All models are trained on eight NVIDIA H100 GPUs with a global batch size of $256$. Appendix~\ref{app:data} details the data construction, and Appendix~\ref{app:implementation} gives the training and inference settings for Single and Dual.

\paragraph{Tasks and protocol}
We evaluate on the real DROID setup across four suites: uncluttered, near-object Pick-and-Place; Pick-and-Place with Distractors; Fold/Unfold with near and far cloth placements; and Move with far objects. All methods share the same $12$ randomized object, placement, and prompt instances per suite ($48$ trials total). The external cameras are placed by hand without extrinsic calibration. Appendix~\ref{app:real_world} lists the complete task instances and success criteria. Trials run at a target $15$\,Hz for at most $900$ control steps, stopping early if continued execution risks robot damage. Success requires full task completion; we report mean success $\pm$ standard error of the mean (SEM), in percent, over $12$ trials per suite and pool all $48$ for Overall.

\paragraph{Models and comparisons}
\textbf{\modelname{}-Single} and \textbf{\modelname{}-Dual} use one and two external RGB-D views, respectively, with full and foveal RGB and depth inputs (four images per view). We compare against wrist-equipped \textbf{\boldmath\pifive}, which shares our base checkpoint and training data, and its deployment-time wrist blackout, \textbf{\boldmath\baselineBlackout}; \textbf{\boldmath\pifive-Single-External} and \textbf{\boldmath\pifive-Dual-External} variants,
which were post-trained with one or two full external RGB views without a wrist view, box prediction, or crops; and \textbf{\boldmath\pifive-Single-External-RGB-D} and \textbf{\boldmath\pifive-Dual-External-RGB-D}, which add depth to the full views. \textbf{\baselineECoT}~\cite{zawalski2024robotic} uses two external RGB views and our box supervision but no crops. \textbf{\ablSingleRGB} and \textbf{\altDualRGB} retain foveation while removing depth from Single and Dual.

The blackout tests how much \pifive relies on the wrist view at deployment, and the external-only baselines test how well a policy learns from external views alone.

\subsection{Replacing the Wrist Camera (Q1)}
\label{sec:exp:real}

\begin{table*}[t]
\caption{Real-world success (\%, mean $\pm$ SEM) on $12$ shared trials per suite; Overall pools all $48$. \textbf{Bold}/\underline{underline}: best/second-best; tied scores share the mark.}
\label{tab:results}
\centering
\small
\setlength{\tabcolsep}{6pt}
\renewcommand{\arraystretch}{1.2}
\resizebox{\textwidth}{!}{%
\begin{tabular}{l cccc c}
\toprule
Method & Pick-and-Place (plain) & Pick-and-Place (w/ Distractors) & Fold/Unfold & Move (far objects) & Overall \\
\midrule
\pifive~\cite{black2025pi_} & \textbf{\val{100.0}{0.0}} & \val{25.0}{13.1} & \val{41.7}{14.9} & \val{41.7}{14.9} & \val{52.1}{7.3} \\
\baselineBlackout & \val{0.0}{0.0} & \val{0.0}{0.0} & \val{0.0}{0.0} & \val{0.0}{0.0} & \val{0.0}{0.0} \\
\pifive-Single-External & \val{16.7}{11.2} & \val{0.0}{0.0} & \val{0.0}{0.0} & \val{8.3}{8.3} & \val{6.3}{3.5} \\
\midrule
\pifive-Dual-External & \val{50.0}{15.1} & \val{16.7}{11.2} & \val{16.7}{11.2} & \val{16.7}{11.2} & \val{25.0}{6.3} \\
\baselineECoT~\cite{zawalski2024robotic} & \underline{\val{66.7}{14.2}} & \val{16.7}{11.2} & \val{16.7}{11.2} & \val{25.0}{13.1} & \val{31.3}{6.8} \\
\midrule
\pifive-Single-External-RGB-D & \val{8.3}{8.3} & \val{8.3}{8.3} & \val{0.0}{0.0} & \val{8.3}{8.3} & \val{6.3}{3.5} \\
\pifive-Dual-External-RGB-D & \val{58.3}{14.9} & \val{41.7}{14.9} & \val{25.0}{13.1} & \val{41.7}{14.9} & \val{41.7}{7.2} \\
\midrule
\ablSingleRGB & \val{33.3}{14.2} & \val{25.0}{13.1} & \val{8.3}{8.3} & \val{16.7}{11.2} & \val{20.8}{5.9} \\
\altDualRGB & \textbf{\val{100.0}{0.0}} & \val{41.7}{14.9} & \underline{\val{58.3}{14.9}} & \val{66.7}{14.2} & \val{66.7}{6.9} \\
\midrule
\textbf{\modelname{}-Single (ours)} & \textbf{\val{100.0}{0.0}} & \textbf{\val{58.3}{14.9}} & \val{50.0}{15.1} & \underline{\val{75.0}{13.1}} & \underline{\val{70.8}{6.6}} \\
\textbf{\modelname{}-Dual (ours)} & \textbf{\val{100.0}{0.0}} & \underline{\val{50.0}{15.1}} & \textbf{\val{75.0}{13.1}} & \textbf{\val{83.3}{11.2}} & \textbf{\val{77.1}{6.1}} \\
\bottomrule
\end{tabular}%
}
\end{table*}

\modelname{}-Single achieves $70.8$\% overall success using one external RGB-D view and \modelname{}-Dual $77.1$\%, compared with $52.1$\% for wrist-equipped \pifive. Both match the baseline's $100.0$\% success on plain Pick-and-Place and achieve higher success on the other three suites (Fig.~\ref{fig:main_task_results}). Table~\ref{tab:results} reports all baselines and ablations.

\begin{figure*}[t]
  \centering
  \includegraphics[width=\textwidth]{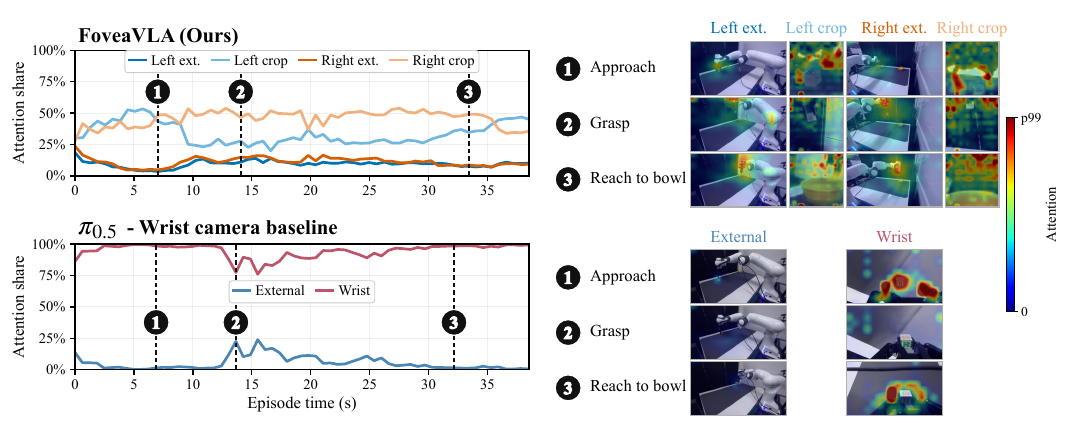}
  \caption{\textbf{Image attention analysis} on a Move episode.
  \pifive~\cite{black2025pi_} attends almost exclusively to the wrist view for the whole episode, while \modelname{} keeps a steady share on the full external views and draws close-up detail from the foveal crops.
  Left: attention share per view over time, with three keyframes marked. Right: attention maps at those keyframes; each model is colored relative to its own 99th percentile, so brightness is not comparable across the two panels.
  }
  \label{fig:attention}
\end{figure*}

\paragraph{Close-up detail across the workspace}
With a single external RGB-D view, success on Distractors rises from the wrist-equipped baseline's $25.0$\% to $58.3$\%, and success on Move rises from $41.7$\% to $75.0$\%. Dual achieves the strongest Fold/Unfold and Move results ($75.0$\% and $83.3$\%). Figure~\ref{fig:attention} suggests why: \pifive places almost all image attention on the wrist view and almost none on the external view, so it can miss a target outside the wrist view. \modelname{} keeps attending to the full external views, which cover the workspace, while the crops add close-up detail.

\paragraph{Inference latency}
On an RTX 5080, \modelname{}-Single takes approximately $125$\,ms per policy query (prefix encoding and action generation) and $50$\,ms for cached-prefix box decoding. \modelname{}-Dual takes roughly twice as long. Box timing includes all camera views and coordinate conversion.
At our control rate, box decoding fits within action-chunk execution, so the predicted boxes are ready when the next observation arrives.

\subsection{Local Detail and Geometric Cues}
\label{sec:exp:ablations}

The ablations in Table~\ref{tab:results} show that replacing the wrist view needs two things: a detailed view of the interaction and geometric cues relating the gripper to its target.

\paragraph{Observing the predicted region (Q2)}
Foveal crops improve success over full views alone, with both RGB and RGB-D. With RGB-D inputs, the single- and dual-view baselines without crops achieve $6.3$\% and $41.7$\% overall, compared with $70.8$\% and $77.1$\% with foveal crops, with gains in all four suites. With two RGB views, we can separate the effect of box supervision from that of the crops: \pifive-Dual-External reaches $25.0$\%, box supervision without crops (\baselineECoT) reaches $31.3$\%, and \altDualRGB reaches $66.7$\%. Box supervision alone adds $6.3$ percentage points, and the crops add a further $35.4$.

\paragraph{Combining detail with depth (Q3)}
With a single view, neither cue suffices on its own. Foveal crops without depth achieve $20.8$\% overall success, and depth without crops achieves $6.3$\%; bringing foveal detail and depth together raises success to $70.8$\%. Depth also improves the dual-view models, from $25.0$\% to $41.7$\% with full views and from $66.7$\% to $77.1$\% with foveation.

\paragraph{A complementary viewpoint (Q3)}
With foveal crops but no depth, a second external view raises success from $20.8$\% to $66.7$\%, approaching the $70.8$\% of Single with RGB-D. This suggests that a second view supplies part of the geometric information that depth provides: adding depth yields a larger gain with one view than with two ($50.0$ versus $10.4$ percentage points). With RGB-D, the second view raises overall success from $70.8$\% to $77.1$\%, with gains on Fold/Unfold and Move, while Single remains best on Distractors ($58.3$\% versus $50.0$\%).

\subsection{Zero-Shot Generalization (Q4)}
\label{sec:exp:molmo}

We deploy the \modelname{}-Dual zero-shot on MolmoSpaces~\cite{kim2026molmospaces}, a photorealistic simulation benchmark with a DROID-matched Franka setup and procedurally generated scenes unseen in training. Our evaluation uses two adapted external camera views and simulator-provided depth; Table~\ref{tab:molmospaces} lists published \pifive-DROID leaderboard scores for reference. Appendix~\ref{app:molmospaces} specifies the observation and evaluation protocols.
\begin{table}[t]
\caption{Zero-shot success (\%) on MolmoSpaces. \pifive-DROID: reported leaderboard reference; our camera and depth protocol is detailed in Appendix~\ref{app:molmospaces}. Average: unweighted task mean. \textbf{Bold}: better score.}
\label{tab:molmospaces}
\centering
\footnotesize
\setlength{\tabcolsep}{8pt}
\renewcommand{\arraystretch}{1.15}
\begin{tabular}{l cc}
\toprule
Task & \pifive-DROID & \modelname{}-Dual (ours) \\
\midrule
Pick-v1.5                & 18.1 & \textbf{33.7} \\
Pick-v2-Classic          & 6.4 & \textbf{9.7} \\
Pick-v2-Filament         & 7.0 & \textbf{10.6} \\
Pick-v2-RandCam          & \textbf{8.0} & 5.5 \\
Pick\&Place-v2           & 12.0 & \textbf{17.0} \\
Pick\&Place-NextTo-v2    & 10.3 & \textbf{14.9} \\
Pick\&Place-Color-v2     & 10.4 & \textbf{20.8} \\
Open-v1                  & 22.7 & \textbf{25.6} \\
Close-v1                 & \textbf{65.1} & 63.9 \\
\midrule
Average                  & 17.8 & \textbf{22.4} \\
\bottomrule
\end{tabular}
\end{table}

Over $8{,}915$ trials, \modelname{}-Dual achieves $22.4$\% average zero-shot success, versus the $17.8$\% published reference, with higher scores on seven of nine tasks. The largest differences are on Pick-v1.5 ($33.7$\% versus $18.1$\%) and Pick\&Place-Color-v2 ($20.8$\% versus $10.4$\%).

\subsection{Foveation Prediction (Q5)}
\label{sec:exp:fovea}

We evaluate how well \modelname{} predicts the next interaction region on held-out DROID interaction boxes, alongside PaliGemma~\cite{beyer2024paligemma}, \pifive-DROID, SAM3~\cite{carion2026sam}, and fine-tuned YOLO11s~\cite{yolo11_ultralytics} (Table~\ref{tab:foveation}).
Our model predicts the region at $t+w$ from frame $t$. PaliGemma, SAM3, and YOLO11s receive the target frame and annotation-derived role information; PaliGemma and SAM3 use role-specific text prompts, while YOLO11s uses the annotated primitive to select the detected roles.
\begin{table}[t]
\caption{Held-out DROID box localization. FT: box fine-tuning; PG: projected gripper ($*$: fitted on training annotations). Wait: added delay after a new observation arrives. \textbf{Bold}/\underline{underline}: best/second-best, including ties.}
\label{tab:foveation}
\centering
\footnotesize
\setlength{\tabcolsep}{3pt}
\renewcommand{\arraystretch}{1.15}
\resizebox{0.9\columnwidth}{!}{%
\begin{tabular}{l cc ccc}
\toprule
Method & FT & Frame & Acc@.5 $\uparrow$ & mIoU $\uparrow$ & Wait (ms) $\downarrow$ \\
\midrule
\pifive-DROID & No & $t$ & 0.002 & 0.009 & \textbf{0} \\
PaliGemma + PG & No$^{*}$ & $t+w$ & 0.132 & 0.207 & 427 \\
SAM3 + PG & No$^{*}$ & $t+w$ & 0.305 & 0.347 & 690 \\
YOLO11s & Yes & $t+w$ & \textbf{0.488} & \textbf{0.492} & \underline{29} \\
\midrule
\textbf{\modelname{} (ours)} & Yes & $t$ & \underline{0.483} & \underline{0.465} & \textbf{0} \\
\bottomrule
\end{tabular}%
}
\end{table}

Our future-box predictions achieve $0.465$ mIoU and $0.483$ Acc@.5, comparable to YOLO11s at $0.492$ and $0.488$ despite predicting before the target frame is available. They also exceed \pifive-DROID, PaliGemma + PG, and SAM3 + PG. The increase from \pifive-DROID's $0.009$ mIoU shows the value of interaction-box supervision. Because \modelname{} decodes its box during action execution, it adds no wait when the new observation arrives, whereas YOLO11s, which needs the target frame, adds $29$\,ms.

\section{Conclusion}

\modelname makes the selection of visual detail part of the robot policy: alongside each action chunk, it predicts the interaction regions that form its next foveal observations. On four real-world task suites, \modelname{}-Single achieves $70.8$\% success with one external RGB-D view, versus $52.1$\% for the wrist-equipped baseline, and \modelname{}-Dual reaches $77.1$\%. With a single view, both foveal detail and depth are needed. \modelname{}-Dual also transfers zero-shot to simulated MolmoSpaces scenes. Because it needs no wrist view, \modelname could also learn from demonstrations recorded without wrist cameras, including human videos.

\section*{Acknowledgment}
This work was supported by NSF:FRR 2153854, the Stanford Institute for Human-Centered AI (HAI), and Amazon. This work used Marlowe~\cite{marlowe2025}, Stanford University's GPU-based Computational Instrument, supported by Stanford HAI and Stanford Research Computing. We thank Michael Piseno, Haoquan Fang, Chongkai Gao, Zhi (Leo) Wang, and Baifeng Shi for their insightful discussions.

\bibliographystyle{IEEEtran}
\bibliography{IEEEabrv,references}

\clearpage
\useRomanappendicesfalse
\appendices
\section{Dataset Construction}
\label{app:data}

\subsection{Matching demonstrations and annotations}
We construct training examples by joining DROID demonstrations with RoboInter's frame-level interaction annotations~\cite{khazatsky2024droid,li2026robointer}. DROID's RLDS records provide the robot state, action sequence, and task instruction; the corresponding raw recordings provide external-camera images at $1280\times720$ resolution and stereo recordings for depth estimation. We match the two by episode, camera view, and frame index, so each crop, box target, and robot action come from the same timestep. We use the rectified left imager of each external stereo camera as the RGB reference. The wrist stream is not used to construct \modelname's observations.

\subsection{Interaction-box construction}
An interaction box encloses the visual components required by the annotated primitive skill. We use three annotation roles: the gripper, the manipulated object or tool, and the target region, which RoboInter labels as a placement proposal. Table~\ref{tab:appendix_roles} gives the role mapping used in preprocessing.

\begin{table}[!htbp]
\caption{Required components for interaction-box construction from RoboInter primitive skills.}
\label{tab:appendix_roles}
\centering
\footnotesize
\setlength{\tabcolsep}{4pt}
\begin{tabular}{@{}p{0.53\columnwidth}p{0.40\columnwidth}@{}}
\toprule
Primitive skills & Required components \\
\midrule
Pick, push, pull, press, twist & Gripper and object/tool \\
Transfer, place, slide, fold, pour, insert, throw, manipulate, shake, strike & Gripper, object/tool, and target region \\
\bottomrule
\end{tabular}
\end{table}

We normalize each required component box by image width and height, discarding boxes with invalid coordinates. If $\mathcal{S}(s_t)$ is the set of roles required by primitive $s_t$ and $c_{r,t}^v$ is the box for role $r$ in view $v$, the target is
\begin{equation}
 b_t^v=\operatorname{BBox}\!\left(\bigcup_{r\in\mathcal{S}(s_t)}c_{r,t}^v\right),
 \label{eq:appendix_box_union}
\end{equation}
where $\operatorname{BBox}$ returns the smallest enclosing axis-aligned rectangle. Frames with a missing or invalid required component, or an unsupported primitive, are excluded. Primitive labels are used to construct supervision and are not provided to \modelname{} at inference.

\subsection{Temporal pairing and sample validity}
A sample at $t$ uses $b_t^v$ to construct its foveal observation and $b_{t+8}^v$ as the prediction target ($w=8$). Both annotations must be valid, and the future frame must exist in the same demonstration. The future box is constructed using the primitive annotated at the future frame, which allows the target to change across a skill transition. We intersect these eligible samples with the DROID non-idle timestep ranges used by the OpenPI training pipeline~\cite{openpi} and with the availability of aligned depth. Single uses the physical-left external view; Dual additionally requires valid inputs and targets for the physical-right view, paired with the left view at the same timestep.

\subsection{Foveal crops and augmentation}
The prediction target is the tight interaction box; the visual encoder receives a larger square crop around it. Let $m$ be the larger of the box's pixel width and height. The nominal crop is centered on the box and has side length $1.5m$. We adjust its position and size at image boundaries while retaining the interaction box. When a square containing the box cannot fit inside the image, the out-of-image area is padded with zeros. The crop is extracted from the $1280\times720$ source image and then resized to $224\times224$. Full views preserve aspect ratio and are padded to the same encoder resolution.

The prepared training crops use a mixture of $45\%$ nominal crops, $45\%$ perturbed crops, and $10\%$ full views. For a perturbed crop, each center coordinate is shifted uniformly by up to $0.25m$, and the nominal side length is multiplied by a log-uniform factor in $[1,4/3]$. The resulting square is constrained to retain the annotated interaction region. The full views match the initial inference query, before any predicted box is available. The prediction target remains the unperturbed future box.

\subsection{Aligned metric depth}
For each external camera, we recover its rectified stereo pair and camera calibration from the raw DROID recording. Fast-FoundationStereo~\cite{wen2025fastfoundationstereo} estimates disparity from each camera's own stereo pair at $640\times360$ resolution, using eight refinement iterations and a maximum disparity of $192$. After accounting for the resize scale, metric optical-axis depth is computed as $z=f_xB/d$, where $f_x$ is the horizontal focal length, $B$ is the stereo baseline in meters, and $d$ is disparity in pixels at the corresponding resolution.

We mark non-finite or non-positive disparity, correspondences outside the right image, and depths outside $[0.2,3.0]$\,m as invalid. The depth map is aligned to the reference RGB image, and the same padding and stored foveal crop coordinates are applied to depth and its validity mask. Real-robot deployment uses the same disparity conversion and crop geometry.

\section{Implementation and Training Details}
\label{app:implementation}

\subsection{Model and optimization}
Single and Dual are separately post-trained from the \pifive base checkpoint using the shared recipe in Table~\ref{tab:appendix_training}. Each retains the SigLIP visual encoder, Gemma language backbone, and flow-matching action expert. RGB images and colorized depth maps use the same visual encoder but occupy separate token sequences. Single supplies four image streams (full and foveal RGB and depth), and Dual supplies eight. We update all model parameters; both the flow and foveation losses propagate into the shared backbone.

\begin{table}[!htbp]
\caption{Training and inference settings for \modelname{}-Single and \modelname{}-Dual.}
\label{tab:appendix_training}
\centering
\footnotesize
\setlength{\tabcolsep}{4pt}
\begin{tabular}{@{}p{0.49\columnwidth}p{0.44\columnwidth}@{}}
\toprule
Setting & Value \\
\midrule
Initialization & \pifive base checkpoint \\
Training updates / global batch & $100{,}000$ / $256$ \\
Training hardware & $8$ NVIDIA H100 GPUs \\
Optimizer & AdamW \\
Adam $(\beta_1,\beta_2)$ / $\epsilon$ & $(0.9,0.95)$ / $10^{-8}$ \\
Weight decay / gradient norm clip & $10^{-10}$ / $1.0$ \\
Learning rate & $5\times10^{-5}$ \\
Warm-up & $1{,}000$ updates \\
Exponential moving average decay & $0.99$ \\
Foveation-loss weight $\lambda$ & $0.1$ \\
Depth-stream dropout & $0.1$ per stream \\
Image / depth-map resolution & $224\times224$ \\
Predicted action horizon $H$ & $16$ control steps \\
Executed horizon / box offset $w$ & $8$ control steps \\
Flow integration steps & $10$ \\
\bottomrule
\end{tabular}
\end{table}

The learning rate warms up to $5\times10^{-5}$ over the first $1{,}000$ updates and remains constant thereafter. We use the DROID normalization statistics supplied with the base checkpoint.

The seven arm-joint action coordinates are represented as offsets from the current joint positions during training; the gripper command remains absolute. State and action normalization follow the base policy's quantile convention, and inverse transformations recover absolute joint-position commands for execution. The physical state and action each have eight dimensions and are padded to the model's $32$-dimensional interface. The normalized robot state is discretized into $256$ bins for the language prefix.

\subsection{Depth representation and visual augmentation}
Depth is encoded on a fixed metric scale shared across all frames. For valid depth $z$, we compute
\begin{equation}
 r(z)=\frac{\log(z/0.2)}{\log(3.0/0.2)}
\end{equation}
and apply a Turbo color map to $1-r(z)$, so near points have warm colors and far points have cool colors. Invalid pixels are black. A $14\times14$ depth patch is masked from attention if fewer than $25\%$ of its pixels are valid. During training, each depth stream is independently dropped with probability $0.1$. RGB receives brightness, contrast, and saturation jitter with magnitudes $0.3$, $0.4$, and $0.5$, respectively. We apply color jitter to RGB only and use no additional geometric augmentation, which would break the alignment between crops, depth, and box coordinates.

\subsection{Box coordinates and decoding}
The predicted box describes the tight interaction region in the full view, before crop expansion. We transform normalized source-image coordinates to the resized and padded $224\times224$ full view, then quantize each coordinate to the nearest of $1{,}024$ bins. The sequence order is $(y_{\min},x_{\min},y_{\max},x_{\max})$, with the physical-left box preceding the physical-right box in Dual. At inference, location-token decoding is greedy. We invert the resize and padding of the full view, clip coordinates to the source-image bounds, and construct the square foveal crop. If a decoded box is invalid, we reuse that view's last valid box, or the full view if none exists.

\subsection{Prediction and execution horizons}
The action expert predicts $H=16$ actions per query, and the controller executes the first $w=8$ before querying again (Sec.~\ref{sec:foveation}). The flow objective in Eq.~(\ref{eq:flow_loss}) supervises the full predicted chunk. Following the base implementation, flow time is sampled as $\tau=0.999u+0.001$ with $u\sim\operatorname{Beta}(1.5,1)$, and inference integrates the flow from noise to actions with ten steps.

\section{MolmoSpaces Evaluation Protocol}
\label{app:molmospaces}

\subsection{Tasks and zero-shot transfer}
We evaluate the \modelname{}-Dual checkpoint without any training on MolmoSpaces data~\cite{kim2026molmospaces}. The nine tasks cover picking, pick-and-place with spatial or color constraints, and opening or closing articulated objects. Task instances, scene initializations, and success definitions follow the official MolmoSpaces benchmarks.

\subsection{External-camera observations}
We render external images at $1280\times720$ so that crops can be extracted at the same source resolution used in training. The first view is fed as the left external camera and the second as the right; full views and predicted foveal crops are resized to $224\times224$. No wrist image is provided.

\begin{table}[!htbp]
\caption{Camera layouts and RGB renderers for our MolmoSpaces evaluation.}
\label{tab:appendix_molmo_cameras}
\centering
\footnotesize
\setlength{\tabcolsep}{3pt}
\begin{tabular}{@{}p{0.37\columnwidth}p{0.17\columnwidth}p{0.38\columnwidth}@{}}
\toprule
Tasks & Renderer & External-camera layout \\
\midrule
Pick-v1.5; Open-v1; Close-v1 & Classic & Original external view plus fixed right-shoulder view \\
Pick-v2-Classic & Classic & Left/right shoulder views \\
Pick-v2-Filament; all three Pick\&Place-v2 tasks & Filament & Left/right shoulder views \\
Pick-v2-RandCam & Filament & Randomized views on opposite sides \\
\bottomrule
\end{tabular}
\end{table}

Table~\ref{tab:appendix_molmo_cameras} summarizes the camera adaptations. For Pick-v1.5, Open-v1, and Close-v1, we retain the original external camera and add a fixed right-shoulder view. For the shoulder-camera tasks, we preserve the recorded left-camera pose and add a right-camera pose with independent position perturbations of up to $5$\,cm and orientation perturbations of up to $8^\circ$ around the mirrored nominal pose. For Pick-v2-RandCam, the two randomized ZED2-like views are constrained to opposite sides of the robot base with a $5$\,cm lateral margin.

\subsection{Depth and action interfaces}
The simulation evaluation supplies metric depth directly from the simulator, aligned with each external RGB image; real-robot deployment estimates depth with Fast-FoundationStereo. The policy receives full and foveal depth using the same crop geometry and fixed metric color encoding as in training.

The policy receives the simulated Franka joint positions, gripper state, and task instruction in DROID format; its joint-position and gripper commands run in simulation.

As on the real robot, eight actions are executed per query before new images are captured and cropped with the predicted boxes. The control period is $66$\,ms. The seven picking and pick-and-place tasks use their benchmark-defined time horizons, while Open-v1 and Close-v1 have a $450$-step limit ($29.7$\,s). Episodes stop early on benchmark success.

\subsection{Scoring and published reference}
The evaluation covers $1{,}000$ trials for each task except Close-v1, whose source benchmark contains $915$ trials. We score episodes with the official MolmoSpaces evaluator, and report the unweighted mean of the nine task success rates.

The \pifive-DROID values in Table~\ref{tab:molmospaces} come from the \href{https://molmospaces.allen.ai/leaderboard}{official MolmoSpaces leaderboard}. Its camera layout, input resolution, and depth availability differ from our evaluation, so these scores serve as a published reference rather than a controlled comparison; the real-world ablations (Sec.~\ref{sec:exp:ablations}) isolate the effects of foveation, depth, and view count.

\section{Real-World Benchmark Details}
\label{app:real_world}

\subsection{Task instances and execution}
\begin{figure*}[p]
  \centering
  \includegraphics[width=0.96\textwidth]{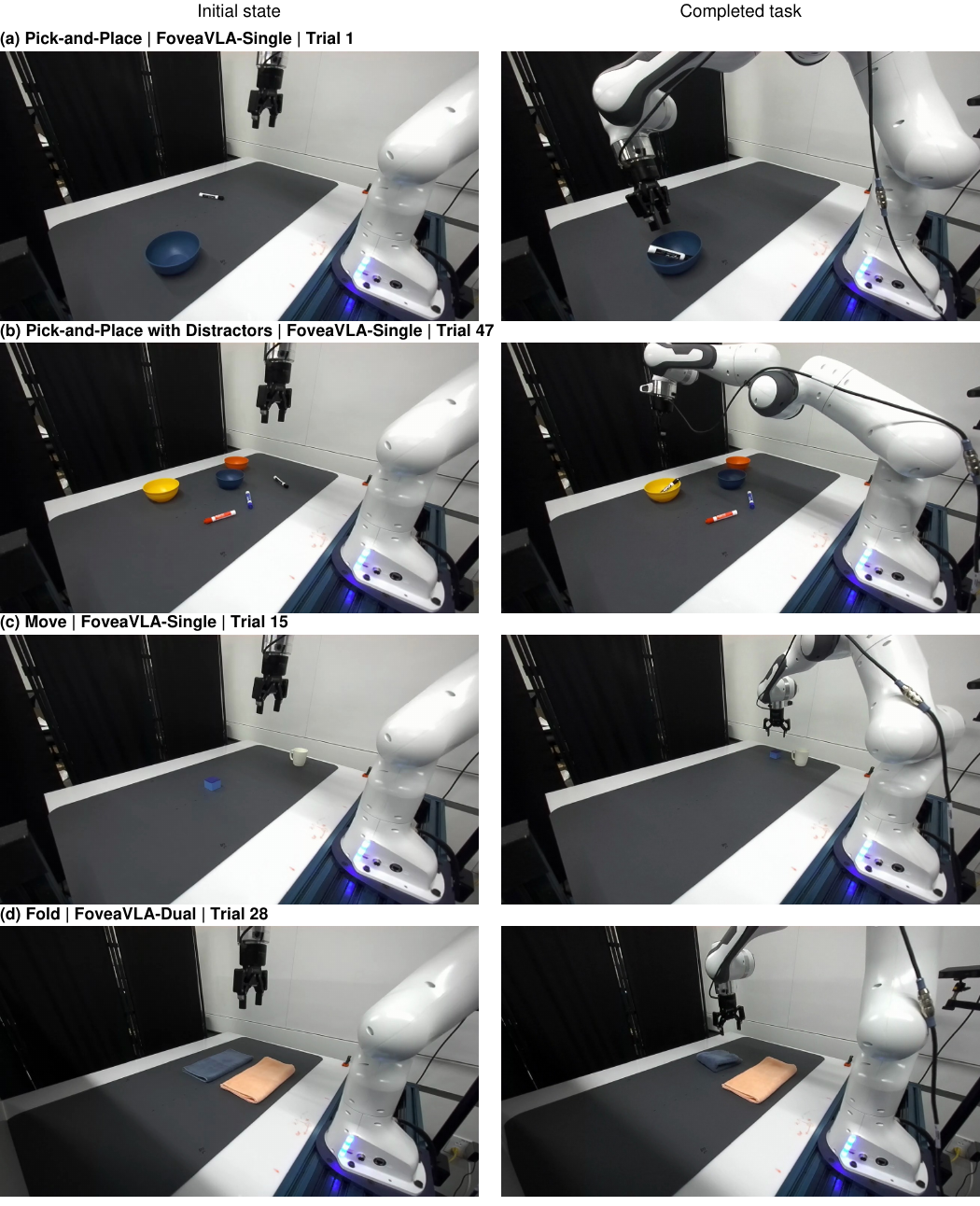}
  \caption{\textbf{Representative real-world task instances.} Initial and completed states for Pick-and-Place, Pick-and-Place with Distractors, Move, and Fold/Unfold, using the trial identifiers in Tables~\ref{tab:real_pick_place}--\ref{tab:real_distractors}. The first three rows show \modelname{}-Single; the last shows \modelname{}-Dual from one of its two external views.}
  \label{fig:real_world_examples}
\end{figure*}

\begin{figure*}[!t]
  \centering
  \includegraphics[width=\textwidth]{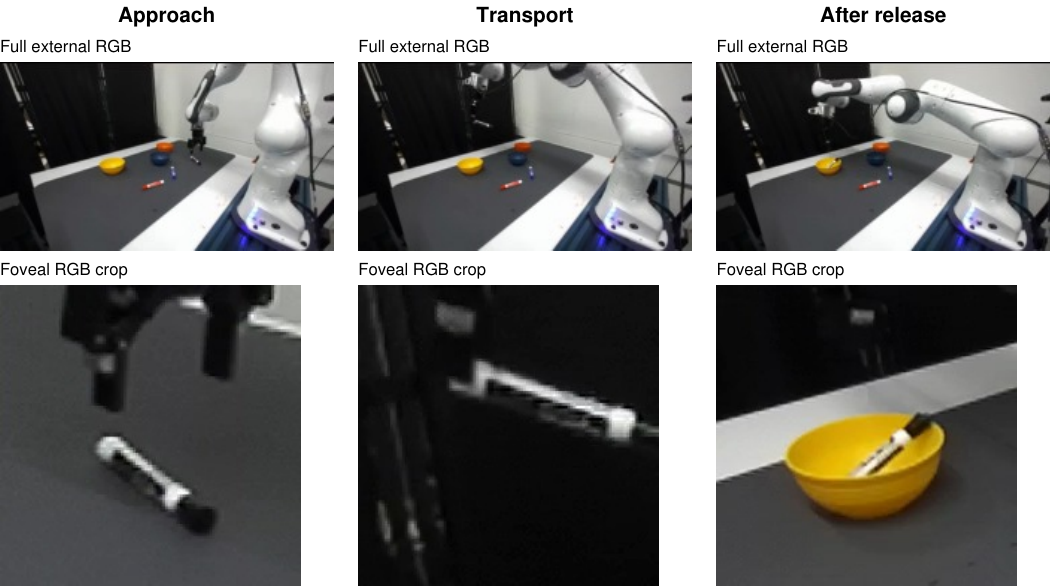}
  \caption{\textbf{Foveal observations during a successful distractor trial.} \modelname{}-Single executes ``Place the black marker in the yellow bowl'' (trial $47$). Columns show three moments in temporal order. Each column shows the full external RGB view and the foveal RGB crop from the same frame (depth omitted). The crops show the target marker during approach and transport, and the marker in the target bowl after release.}
  \label{fig:real_foveation_sequence}
\end{figure*}

The real-world benchmark contains four task suites with $12$ instances each. An instance specifies the objects, their initial placement, and a language instruction; the same instances are used across methods. Our tasks adapt the tabletop tasks of Cloak~\cite{piseno2026cloak}, with changes to the object set and towel setups and a Pick-and-Place with Distractors suite in place of Cloak's Remove task. Tables~\ref{tab:real_pick_place}--\ref{tab:real_distractors} list all $48$ instances. Figure~\ref{fig:real_world_examples} shows representative initial and completed states from all four suites. Figure~\ref{fig:real_foveation_sequence} illustrates the full views and foveal crops during a successful distractor trial.

We use the DROID robot setup described in Sec.~\ref{sec:exp:setup}. Each trial runs at a target $15$\,Hz for at most $900$ control steps and can be stopped earlier to avoid robot damage. The object lists and positions are used only to set up the scene and are not given to the policy. Positions are named tabletop slots: FL, FR, BL, and BR denote front-left, front-right, back-left, and back-right, respectively, and FM and BM denote the middle position in the front and back rows.

\subsection{Task definitions and success criteria}
\paragraph{Pick-and-Place}
The policy must move the specified source object into the specified container. The instances vary the source among markers, cubes, and towels and the container among bowls, a box, and a large pot. Success requires grasping the specified source and releasing it inside the correct container.

\paragraph{Move}
The policy must move the source object next to a reference object. Success requires grasping the source and releasing it within one source-object width of the reference without moving the reference. Instances use cubes, cups, and towels as sources and cups, bowls, and a large pot as references.

\paragraph{Fold/Unfold}
The $12$ trials comprise four groups of three: folding one towel, folding a target towel with a distractor present, unfolding one towel, and unfolding a target towel with a distractor present. Each group targets blue, grey, and pink towels once each. For folding, every towel in the scene starts flat; for unfolding, every towel starts with one half-fold. Thus, in the two-towel scenes, initial fold state does not reveal the target. The instruction specifies the target color. Folding requires one half-fold with opposite edges aligned; unfolding requires a flat towel with no folded-over section. Success requires completing the requested manipulation on the specified towel and releasing it.

\paragraph{Pick-and-Place with Distractors}
The four scene compositions are two markers with one bowl, two markers with two bowls, three markers with two bowls, and three markers with three bowls, with three trials per composition. Markers occupy the back row and empty bowls the front row. A two-object row uses its left and right positions; a three-object row additionally uses the middle position, which is occupied by a distractor. A single bowl is placed at front-middle. The instruction names both the marker and bowl colors, and each group uses all three marker colors and all three bowl colors once as targets. Success requires placing the instructed marker inside the instructed bowl.

\subsection{Outcome annotation}
An operator assigns binary success after each attempt according to the task-specific criteria above. In scenes with distractors, disturbing a distractor is allowed as long as the instructed manipulation is completed. Per-suite success is averaged over its $12$ trials, and Overall pools all $48$ trials, with SEM reported as in the main text.

\subsection{Complete trial specifications}
The tables list the exact instructions given to the policy, which sometimes use alternative words such as ``pen'' for a marker and ``cloth'' for a towel. Within the Fold/Unfold and Distractors suites, only the objects listed for an instance are placed on the table. The Fold/Unfold initial-state rules above apply to both the target and any distractor towel.

\begin{table*}[p]
\caption{Pick-and-Place trial specifications. S denotes the source object and T the target container.}
\label{tab:real_pick_place}
\centering
\footnotesize
\setlength{\tabcolsep}{4pt}
\renewcommand{\arraystretch}{1.25}
\begin{tabular}{@{}p{0.03\textwidth}p{0.51\textwidth}p{0.39\textwidth}@{}}
\toprule
Trial & Objects and initial positions & Language instruction \\
\midrule
0 & S: red marker (BR); T: blue bowl (FR) & Place the marker in the blue bowl \\
1 & S: black marker (FL); T: blue bowl (BL) & Put the pen in the bowl \\
2 & S: red marker (BL); T: orange bowl (FR) & Place the red marker in the orange bowl \\
3 & S: small cube (FL); T: orange bowl (FR) & Place the small block in the bowl \\
4 & S: blue cube (FR); T: blue bowl (BL) & Place the block in the bowl \\
5 & S: small cube (BR); T: blue bowl (FR) & Place the small block in the blue bowl \\
6 & S: blue cube (FL); T: box (FR) & Put the blue block in the box \\
7 & S: small cube (BR); T: box (BL) & Place the small cube in the box \\
8 & S: small cube (BL); T: box (FR) & Put the small cube in the box \\
9 & S: grey towel (BL); T: large pot (FR) & Place the grey towel in the large pot \\
10 & S: pink towel (FL); T: large pot (BR) & Put the towel in the large pot \\
11 & S: blue towel (FR); T: large pot (FL) & Put the blue cloth in the large pot \\
\bottomrule
\end{tabular}
\end{table*}

\begin{table*}[p]
\caption{Move trial specifications. S denotes the source object and R the reference object.}
\label{tab:real_move}
\centering
\footnotesize
\setlength{\tabcolsep}{4pt}
\renewcommand{\arraystretch}{1.25}
\begin{tabular}{@{}p{0.03\textwidth}p{0.51\textwidth}p{0.39\textwidth}@{}}
\toprule
Trial & Objects and initial positions & Language instruction \\
\midrule
12 & S: blue cube (BL); R: green cup (BR) & Move the cube next to the cup \\
13 & S: small cube (FR); R: white cup (FL) & Move the cube next to the white mug \\
14 & S: blue cube (BR); R: silver cup (BL) & Move the block next to the cup \\
15 & S: blue cube (FL); R: white cup (BR) & Move the blue block near the white mug \\
16 & S: silver cup (BL); R: yellow bowl (BR) & Move the cup near the yellow bowl \\
17 & S: silver cup (FL); R: orange bowl (BL) & Move the silver mug near the bowl \\
18 & S: white cup (FR); R: yellow bowl (FL) & Move the white mug near the bowl \\
19 & S: white cup (FR); R: blue bowl (BL) & Move the mug next to the bowl \\
20 & S: pink towel (FL); R: large pot (BR) & Move the pink towel next to the large pot \\
21 & S: grey towel (BL); R: large pot (FR) & Move the grey towel near the large pot \\
22 & S: blue towel (FR); R: large pot (FL) & Move the towel near the large pot \\
23 & S: blue towel (FL); R: large pot (FR) & Move the blue towel next to the large pot \\
\bottomrule
\end{tabular}
\end{table*}

\begin{table*}[p]
\caption{Fold/Unfold trial specifications. T denotes the target towel and D the distractor. Fold trials start flat; unfold trials start with one half-fold.}
\label{tab:real_fold}
\centering
\footnotesize
\setlength{\tabcolsep}{4pt}
\renewcommand{\arraystretch}{1.25}
\begin{tabular}{@{}p{0.03\textwidth}p{0.51\textwidth}p{0.39\textwidth}@{}}
\toprule
Trial & Objects and initial positions & Language instruction \\
\midrule
24 & T: blue towel (FL) & Fold the blue towel \\
25 & T: grey towel (FR) & Fold the grey towel \\
26 & T: pink towel (BL) & Fold the pink towel \\
27 & T: blue towel (FL); D: grey towel (BL) & Fold the blue towel \\
28 & T: grey towel (FR); D: pink towel (BR) & Fold the grey towel \\
29 & T: pink towel (BL); D: blue towel (FL) & Fold the pink towel \\
30 & T: blue towel (FL) & Unfold the blue towel \\
31 & T: grey towel (FR) & Unfold the grey towel \\
32 & T: pink towel (BL) & Unfold the pink towel \\
33 & T: blue towel (FL); D: grey towel (BL) & Unfold the blue towel \\
34 & T: grey towel (FR); D: pink towel (BR) & Unfold the grey towel \\
35 & T: pink towel (BL); D: blue towel (FL) & Unfold the pink towel \\
\bottomrule
\end{tabular}
\end{table*}

\begin{table*}[p]
\caption{Pick-and-Place with Distractors trial specifications. S denotes the target marker, T the target bowl, and D all distractors.}
\label{tab:real_distractors}
\centering
\footnotesize
\setlength{\tabcolsep}{4pt}
\renewcommand{\arraystretch}{1.25}
\begin{tabular}{@{}p{0.03\textwidth}p{0.51\textwidth}p{0.39\textwidth}@{}}
\toprule
Trial & Objects and initial positions & Language instruction \\
\midrule
36 & S: red marker (BL); T: blue bowl (FM); D: blue marker (BR) & Place the red marker in the blue bowl \\
37 & S: blue marker (BR); T: orange bowl (FM); D: black marker (BL) & Place the blue marker in the orange bowl \\
38 & S: black marker (BL); T: yellow bowl (FM); D: red marker (BR) & Place the black marker in the yellow bowl \\
39 & S: red marker (BR); T: blue bowl (FL); D: blue marker (BL); orange bowl (FR) & Place the red marker in the blue bowl \\
40 & S: blue marker (BL); T: orange bowl (FR); D: black marker (BR); yellow bowl (FL) & Place the blue marker in the orange bowl \\
41 & S: black marker (BR); T: yellow bowl (FL); D: red marker (BL); blue bowl (FR) & Place the black marker in the yellow bowl \\
42 & S: red marker (BL); T: blue bowl (FR); D: blue marker (BM); black marker (BR); orange bowl (FL) & Place the red marker in the blue bowl \\
43 & S: blue marker (BR); T: orange bowl (FL); D: black marker (BL); red marker (BM); yellow bowl (FR) & Place the blue marker in the orange bowl \\
44 & S: black marker (BL); T: yellow bowl (FR); D: red marker (BM); blue marker (BR); blue bowl (FL) & Place the black marker in the yellow bowl \\
45 & S: red marker (BR); T: blue bowl (FL); D: blue marker (BL); black marker (BM); orange bowl (FM); yellow bowl (FR) & Place the red marker in the blue bowl \\
46 & S: blue marker (BL); T: orange bowl (FR); D: black marker (BM); red marker (BR); yellow bowl (FL); blue bowl (FM) & Place the blue marker in the orange bowl \\
47 & S: black marker (BR); T: yellow bowl (FL); D: red marker (BL); blue marker (BM); blue bowl (FM); orange bowl (FR) & Place the black marker in the yellow bowl \\
\bottomrule
\end{tabular}
\end{table*}

\end{document}